\documentclass[runningheads,a4paper]{llncs}

\usepackage{amsmath}
\usepackage{amssymb}
\usepackage{graphicx}
\usepackage{pgfplots}
\pgfplotsset{compat=1.5}
\usepackage{subfigure}
\usepackage{subfig}
\usepackage{mwe}
\usepackage{algorithm}
\usepackage{algpseudocode}
\usepackage{float}

\usepackage{url}
\urldef{\mailsa}\path|markku.hinkka@aalto.fi|
\urldef{\mailsb}\path|teemu.lehto@qpr.com|    
\urldef{\mailsc}\path|keijo.heljanko@aalto.fi|
\urldef{\mailsd}\path|alex.jung@aalto.fi|
\newcommand{\keywords}[1]{\par\addvspace\baselineskip
	\noindent\keywordname\enspace\ignorespaces#1}

\begin{document}
	
\mainmatter  

\title{Classifying Process Instances Using Recurrent Neural Networks}

\titlerunning{Classifying Process Instances Using Recurrent Neural Networks}

%
%
\author{Markku Hinkka\inst{1,2} \and Teemu Lehto\inst{1,2} \and Keijo Heljanko\inst{1,3} \and Alexander Jung\inst{1}}
\authorrunning{Markku Hinkka, Teemu Lehto, Keijo Heljanko, Alexander Jung}

\institute{Aalto University, School of Science, Department of Computer Science, Finland
	\and
	QPR Software Plc, Finland \\
	\and
	HIIT Helsinki Institute for Information Technology
	\\
	\mailsa, \mailsb, \mailsc, \mailsd\\
}

%
%

\toctitle{Classifying Process Instances Using Recurrent Neural Networks}
\tocauthor{Markku Hinkka, Teemu Lehto, Keijo Heljanko, Alexander Jung}
\maketitle

\begin{abstract}
Process Mining consists of techniques where logs created by operative systems are transformed into process models. In process mining tools it is often desired to be able to classify ongoing process instances, e.g., to predict how long the process will still require to complete, or to classify process instances to different classes based only on the activities that have occurred in the process instance thus far. Recurrent neural networks and its subclasses, such as Gated Recurrent Unit (GRU) and Long Short-Term Memory (LSTM), have been demonstrated to be able to learn relevant temporal features for subsequent classification tasks. In this paper we apply recurrent neural networks to classifying process instances. The proposed model is trained in a supervised fashion using labeled process instances extracted from event log traces. This is the first time we know of GRU having been used in classifying business process instances. Our main experimental results shows that GRU outperforms LSTM remarkably in training time while giving almost identical accuracies to LSTM models. Additional contributions of our paper are improving the classification model training time by filtering infrequent activities, which is a technique commonly used, e.g., in Natural Language Processing (NLP).

\end{abstract}

\section{Introduction}
\label{introduction}


Unstructured \textit{event logs} generated by systems in business processes are used in Process Mining to automatically build real-life process definitions and as-is models behind those event logs. There are growing number of applications for predicting the properties of newly added event log cases, or process instances, based on case data imported earlier into the system~\cite{evermann2017predicting}\cite{DBLP:journals/corr/Francescomarino15}\cite{DBLP:conf/ssci/NavarinVPS17}\cite{DBLP:conf/caise/TaxVRD17}. The more the users start to understand their own processes, the more they want to optimize them. This optimization can be facilitated by performing predictions. In order to be able to predict properties of the new and ongoing cases, as much information as possible should be collected that is related to the event log traces and relevant to the properties to be predicted. Based on this information, a model of the system creating the event logs can be created. In our approach, the model creation is performed using supervised machine learning techniques.


In paper~\cite{DBLP:conf/bpm/HinkkaLHJ17} we explored the possibility to use machine learning techniques for performing classification and root cause analysis for a process mining related classification task. In the paper, we tested the efficiency of several feature selection techniques and sets of features based on process mining models in the context of a classification task. One of the biggest problems with that approach is that, due to the simplicity of the features that are just numeric values, the user still needs to select and generate the set of features from which to select the final subset of features used for classification. For this purpose, we use Natural Language Processing techniques together with recurrent neural network techniques such as Long Short-Term Memory (LSTM) and Gated Recurrent Unit (GRU), the latter of which has not been used in process mining context before. These techniques can also learn more complicated causal relationships between the features in \textit{activity sequences} in event logs. We have tested several different approaches and parameters for the recurrent neural network techniques and have compared the results with the results we collected in our earlier paper. As in our previous paper, we focus on classification tasks yielding boolean results which can be seen as responding to a query: Does this trace have the selected properties or not? This approach provides a very flexible basis for implementing additional functionalities such as, e.g., predicting the eventual duration, category, or resource usage required for the trace to complete.

The primary motivation for this paper is the need to perform prediction and classification based on \textit{activity sequences} in event logs as accurately as possible and while simultaneously maximizing the throughput. This motivation comes from the need to build a system that can perform classification and prediction activities accurately on user configurable phenomena based on huge event logs collected and analyzed, e.g., using \textit{Big Data processing frameworks} and methods such as those discussed in our earlier paper~\cite{hinkka2016assessing}. Again, we focus on classification response times by targeting web browser based interactive process mining tool where user wants to perform classifications and expects classification results to be shown within a couple of seconds. Due to this requirement, we also performed some additional experiments for a couple of techniques in order to speed up the classification process: Filtering out infrequent activities and truncating repeated infrequent activities.

Based on a the number of released papers on the subject of predictions and process mining, the interest in combining these subjects has been rapidly increasing. However, only in very recent years, deep learning techniques have been used to perform the actual process mining prediction tasks. ~\cite{evermann2017predicting} and ~\cite{DBLP:conf/caise/TaxVRD17} describe techniques for prediction cycle times and next activities of ongoing traces using LSTM. In~\cite{DBLP:conf/ssci/NavarinVPS17}, the authors further improve the LSTM based prediction technique by also incorporating a mechanism for including attributes associated to events. In contrast, our experimental system is designed to be used as a foundation to solve any classification problem based on \textit{activity sequences}, including the prediction of the next activity or cycle time using either LSTM or GRU. In~\cite{DBLP:journals/corr/Francescomarino15}, the authors present a framework for predicting outcomes of user specified predicates for running cases using clustering based on control flow similarities and then performing classification using attributes associated to events. This two phased process and the usage of event attributes are their biggest difference to our one phase process using only activity sequences.

The rest of this paper is structured as follows: Section~\ref{problemsetup} introduces main concepts related to this paper as well as our goals. Section~\ref{testsetup} presents the test system, framework, and the data sets used in implementing our experiments. The results of these experiments will be presented in Section~\ref{testresults}. Finally Section~\ref{conclusions} draws the conclusions from the test results.

\section{Problem Setup}
\label{problemsetup}

The concepts and terminology used throughout this paper mostly follow those commonly used in process mining and machine learning communities. However, the following subsections will provide short introduction to the most important concepts related to this paper. For more detailed examples and discussion about \textit{event logs}, \textit{activity sequences} and other related terminology, see, e.g., the book by van der Aalst~\cite{van2011process}.

\subsection{Classification and Prediction}
\label{classificationandprediction}

A common \textit{supervised analysis task} solved by \textit{machine learning} techniques is to \textit{predict} or \textit{classify} \textit{data points} based on their properties. The properties of data points are often called also as \textit{predictors} or \textit{features}. Predictors as well as \textit{outcomes} of the prediction can be either continuous or they may be of enumerated types which often are also known as \textit{categorical values} or \textit{labels}. When the outcome of a prediction is this kind of a categorical value, such as a binary value, the performed analysis task is often called \textit{classification}. 

Usually classification in machine learning consists of two phases: \textit{training} a \textit{model} and performing the actual predictions using the trained model. In the model training phase, a supervised machine learning algorithm is used to create a model which produces a predicted outcome for a data point given in a form of predictors. This model building is performed by repeatedly feeding the algorithm with training data points consisting of predictors of an actual data point and actual \textit{outcomes} that the modeled system produced for that data point. Eventually the model \textit{learns} to simulate the system it is modeling by becoming better and better in predicting the outcomes for the training set. As a good \textit{training data set} is a representative sample of the actual \textit{test data} to be used on the model, the trained model will also be able to predict also the outcomes of the actual test data. If the accuracy of predictions for a training data is much better than the accuracy achieved for the test data, the model is said to be \textit{overfitting}: The model has been trained with the biases in the training data and it is not able to \textit{generalize} its predictions for test data.

\subsection{Recurrent Neural Networks}
\label{rnn}

\textit{Artificial Neural Networks} are computing systems inspired by biological neural networks constituting animal brains. They consist of simple interconnected units, also known as \textit{neurons}. The whole network can be trained to provide desired outputs for desired inputs. \textit{Recurrent Neural Networks} (RNN) are a kind of \textit{deep neural networks} that are connected in a way that provides the neural network a capability to remember earlier inputs fed into the network or when producing text, the network is capable of remembering what it has produced before. For example, recurrent neural networks can be used to train to produce text sentences. In this case it is essential to know what words have been produced before. RNNs have been used for large variety of problems, such as speech recognition, machine translation and automatic image captioning. Traditional RNNs have an inherent problem called \textit{vanishing gradient problem} that makes it very hard for them to learn long distance dependencies~\cite{DBLP:journals/corr/ChungGCB14}.

\subsubsection{Long Short-Term Memory and Gated Recurrent Unit}
\label{lstm}

To overcome vanishing gradient problem, more complicated cell types have been developed, such as Long Short-Term Memory (LSTM)~\cite{DBLP:journals/neco/HochreiterS97} and Gated Recurrent Units (GRU)~\cite{DBLP:conf/ssst/ChoMBB14}.

Both GRU and LSTM solve the problem using a \textit{gating mechanism} that has multiple layers of gates, which are actually a \textit{layers of neurons}, that optionally let information through. In LSTM, the purposes of the gates are: Forget gate determines what information to throw away from the current hidden state, input gate layer decides which values to update and output gate decides which values the cell should output. In GRU, there are only two gates: Update gate determines how much of the previous hidden state needs to be passed along to the future, whereas reset gate determines how much of the previous hidden state to forget. 

All of these layers are trained as any other neural network which usually involves defining a cost function and using some method of gradient descent to find out the optimal parametrization. The size of the hidden state defines how long vector of numeric values is used to store the internal state of the unit. The larger the hidden state size is, the more the model has potential for learning while also taking more time to train. When using too large hidden states compared to the actual modeled phenomenon, there is also a risk of overfitting the training data.

According to the empirical evaluations~\cite{DBLP:conf/icml/JozefowiczZS15}\cite{DBLP:journals/corr/ChungGCB14}, there is no clear winner on whether GRU or LSTM is the preferred choice. Both the architectures yield models with similar performance characteristics. However, due to GRU having fewer parameters to train, it has the reputation of being somewhat faster to train. In this paper we want to see if these observations carry over to process mining.

\subsection{Natural Language Processing}
\label{nlp}

Natural Language Processing is a field of study that focuses on studying interactions between human languages and computers. It is used to tackle problems involving speech recognition, understanding natural language and generating natural language. In the context of this paper, we use similar approach often used in tasks requiring understanding of a natural language. In a way, we produce a new language that consists of sentences consisting of words that represent activities within traces. Using these artificial sentences and labels, representing the desired labeling of the trace attached to each sentence, we train a deep neural network model to predict the eventual labels for these sentences, even for activity sequences that represent process instances that have not yet finished.

\subsection{Process Instance Classification}
\label{problemformalization}

The goal of this paper was to produce a classification label using a trained RNN for any given activity sequence based only on the activity identifiers contained in the sequence itself. The actual labels for activity sequences used in this paper were of boolean-type, but the used algorithms should work equally well also with more than two possible labels. We did not set any limitations for the actual property being labeled. However, in this paper we concentrated especially in trace throughput time related properties, but it can as easily be related to, e.g., used resources, trace value or its type. We also experimented with a couple of RNN -based approaches in predicting the eventual classification for unfinished traces.

\section{Experimental Setup}
\label{testsetup}

Tests were performed using five publicly available data sets. Table~\ref{table:testdatasets} shows the details of each tested dataset including the number of traces, number of positive classifications, the maximum activity sequence lengths of traces and the number of unique activities. For all the other datasets except BPIC14, we used all the available rows. For BPIC14 we used 40000 first cases of all the available 466616 traces in order for the results to be comparable with our earlier work in~\cite{DBLP:conf/bpm/HinkkaLHJ17}, which had this limitation. For every data set, we selected at least one property that somehow split the model into two segments with roughly 20\%-40\% of all the traces in the positive segment and the rest in the negative. For BPIC14 model we used two boolean labellings: Is the total duration of the case longer than 7 days, and does the case represent a "request for information" or something else. Case duration-based labeling relies only on the contents of the events in the event log, whereas the categorization uses a separate case attribute. For all the other data sets we decided to test only case durations in order for the results to be comparable with the tests performed in~\cite{DBLP:conf/bpm/HinkkaLHJ17} and its extended version ~\cite{DBLP:journals/corr/abs-1710-02823}. In BPIC12 and BPIC13, the duration threshold was set to 2 weeks. In BPIC17, this threshold was set to 4 weeks and in Hospital data set to 20 weeks. 
 
\begin{table*}[!t]
	\centering
	\begin{tabular}{lccccc}
		\hline
		Event Log & \# Traces & \# Positive & \% Positive & Seq. Length & \# Activities \\ \hline
		BPIC14-40k~\cite{https://doi.org/10.4121/uuid:c3e5d162-0cfd-4bb0-bd82-af5268819c35} & 40000 & 8108 / 7473 & 20\% / 19\% & 179 & 39 \\
		BPIC12~\cite{https://doi.org/10.4121/uuid:3926db30-f712-4394-aebc-75976070e91f} & 13087 & 3330 & 25\% & 176 & 36 \\
		BPIC13, incidents~\cite{https://doi.org/10.4121/uuid:500573e6-accc-4b0c-9576-aa5468b10cee} & 7554 & 1579 & 21\% & 124 & 12 \\
		BPIC17~\cite{https://doi.org/10.4121/uuid:5f3067df-f10b-45da-b98b-86ae4c7a310b} & 31509 & 11584 & 37\% & 181 & 26 \\
		Hospital~\cite{https://doi.org/10.4121/uuid:d9769f3d-0ab0-4fb8-803b-0d1120ffcf54} & 1143 & 372 & 33\% & 1201 & 624 \\
		\\
	\end{tabular}
	\caption{Used Event logs and their relevant statistics}
	\label{table:testdatasets}
\end{table*}

The input given to the test framework was a CSV file that was formatted in such a way that every row in the file had one column for the labeling and another column for the activity sequence of a single trace in the source data set. These CSV files were created using QPR ProcessAnalyzer Excel Client-process mining tool\footnote{https://www.qpr.com/products/qpr-processanalyzer}. The used CSV files are available in support materials~\cite{supportmaterials}.

After reading these CSV files into memory, we used standard Natural Language Processing techniques. I.e., every activity sequence is treated as a sentence and every activity identifier as a word in a sentence. These sentences are then converted by assigning a unique integer identifier for each unique activity identifier and also for each classification label. Finally, when sending the activity sequences into the RNN, both in the training and in the actual validation phase, these integers representing activities were "one-hot" encoded. The actual "one-hot" encoded classification label for the trace was used as the expected classification label in the training phase.

In order to enhance the training time performance, we experimented with limiting the number of activity identifiers by only accepting N most common activity identifiers in the training set and using a special \textit{unknown} activity identifier to represent all the rest of the activity identifiers. We also ran an experiment applying an additional truncation step where all continuous sequences of these \textit{unknown} activity identifiers were replaced with just one occurrence of the said activity identifier. 

Testing was performed on a single system having Windows 10 operating system. The used hardware consisted of 3.5 GHz Intel Core i5-6600K CPU with 32 GB of main memory and NVIDIA GeForce GTX 960 GPU having 4 GB of memory. The testing framework was built on the test system using Python programming language. The actual recurrent neural networks were built using Lasagne~\footnote{https://lasagne.readthedocs.io/} library that works on top of Theano~\footnote{http://deeplearning.net/software/theano/} which is an efficient mathematical expression evaluation library that can transparently perform computations in GPU and can also perform symbolic differentiation efficiently. Theano was configured to use GPU via CUDA for expression evaluation. The framework allowed testing several different hyperparameter combinations. The source code of the testing framework is available in support materials~\cite{supportmaterials}.

The model was trained using Adam -gradient descent optimizer that has been found performing well with various types of neural networks~\cite{DBLP:journals/corr/KingmaB14}. We also used fixed learning rate through all the test runs referred to in this paper. Cross-entropy between the predicted and true labeling is used as the model training cost function. Gradient clipping was also used to avoid exploding gradients problem. All the training and prediction was performed in batches of configurable size by creating a batch of sentences and then sending these batches as the training or test data for the RNN to process. Batching is used to improve the efficiency since it enables Theano to distribute calculations in bigger chunks to GPU for parallel processing. All RNN unit gates, nonlinearities and weight matrices were initialized with the default initialization values built-in to Lasagne library. 

In most of the test runs, model was trained for 50 test iterations, each consisting of total of 100000 traces, which translated roughly to minimum of 166 and maximum of 5834 \textit{epochs}, depending on the used dataset. After every iteration, prediction accuracies, Area Under the Receiver Operating Characteristic curves (AUROC) and confusion matrices were calculated for the whole validation data set. The accuracy prediction was performed separately for traces of length 25\%, 50\%, 75\% and 100\% from the original length so that a continuous subsequence starting from the first activity is used. Every test iteration consisted of one hundred thousand training runs to train the model. One training run consisted of one activity sequence and its associated outcome. 

\section{Experimental Results}
\label{testresults}

\begin{figure}[!htbp]
	\begin{minipage}{\linewidth}
		\centering
		\begin{minipage}{0.45\linewidth}
			\begin{figure}[H]
				\includegraphics[trim={2cm 10.2cm 2cm 11.4cm},clip,width=0.9\linewidth]{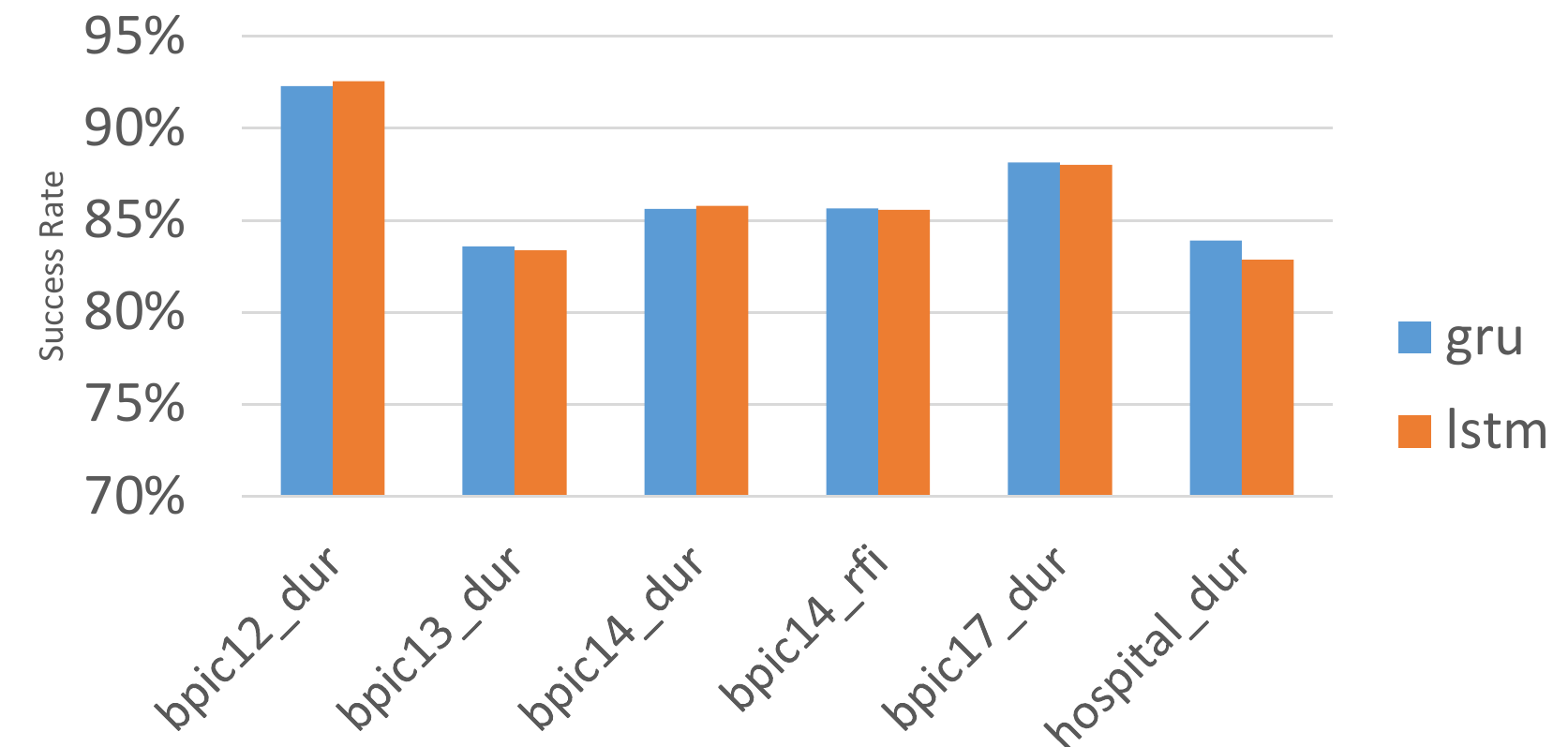}
				\caption[]{Maximum classification accuracy for data sets for both the experimented RNN types}
				\label{figure:accuracy-rnntypes}
			\end{figure}
		\end{minipage}
		\hspace{0.05\linewidth}
		\begin{minipage}{0.45\linewidth}
			\begin{figure}[H]
				\includegraphics[trim={2cm 10.2cm 2cm 11.4cm},clip,width=0.9\linewidth]{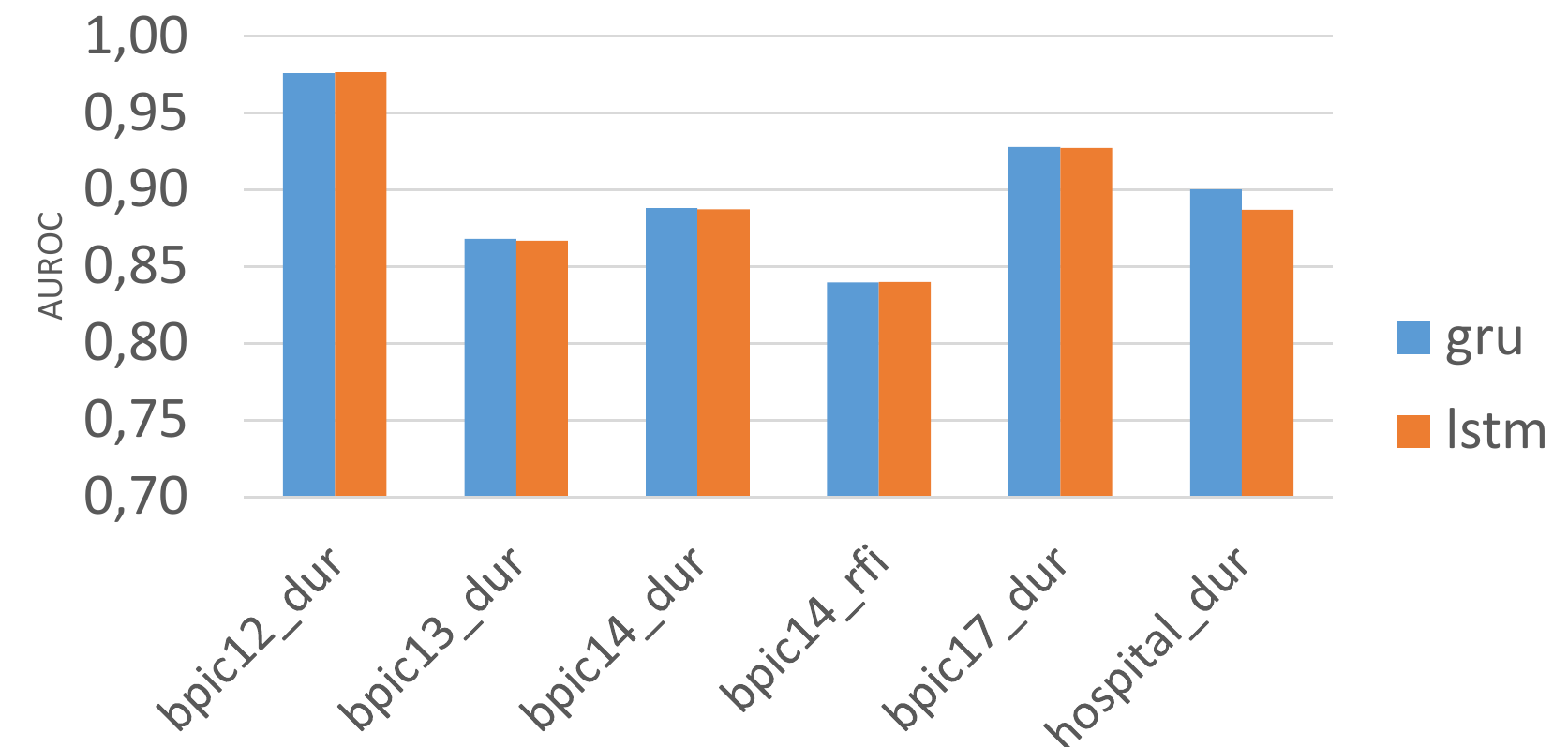}
				\caption[]{Maximum AUROC values for data sets for both the experimented RNN types}
				\label{figure:auroc-rnntypes}
			\end{figure}
		\end{minipage}
	\end{minipage}
	\begin{minipage}{\linewidth}
		\centering
		\begin{minipage}{0.45\linewidth}
			\begin{figure}[H]
				\includegraphics[trim={2cm 10.2cm 2cm 11.4cm},clip,width=0.9\linewidth]{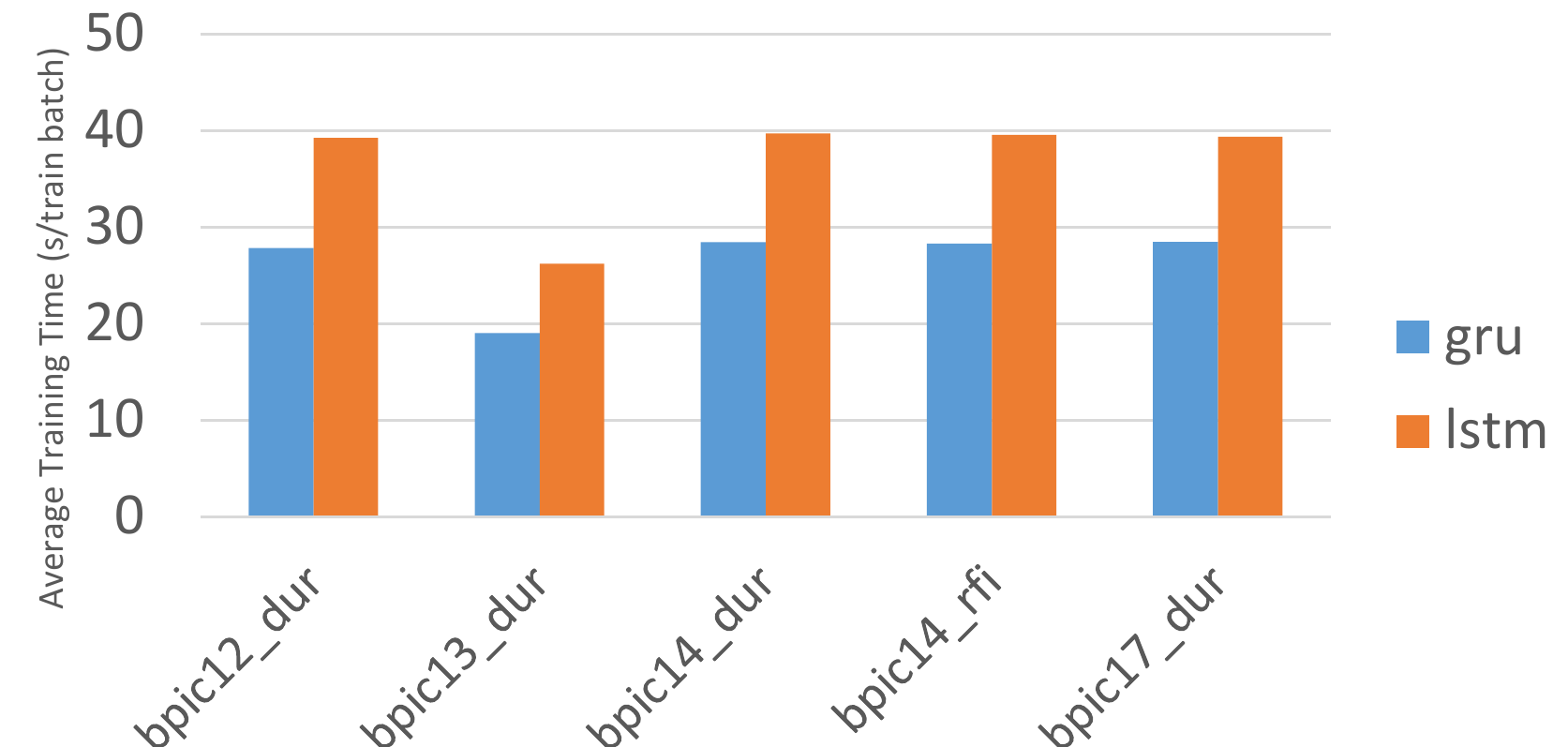}
				\caption[]{Training time usage by RNN type}
				\label{figure:timeusage-rnntypes}
			\end{figure}
		\end{minipage}
		\hspace{0.05\linewidth}
		\begin{minipage}{0.45\linewidth}
			\begin{figure}[H]
				\includegraphics[trim={2cm 10.2cm 2cm 11.4cm},clip,width=0.9\linewidth]{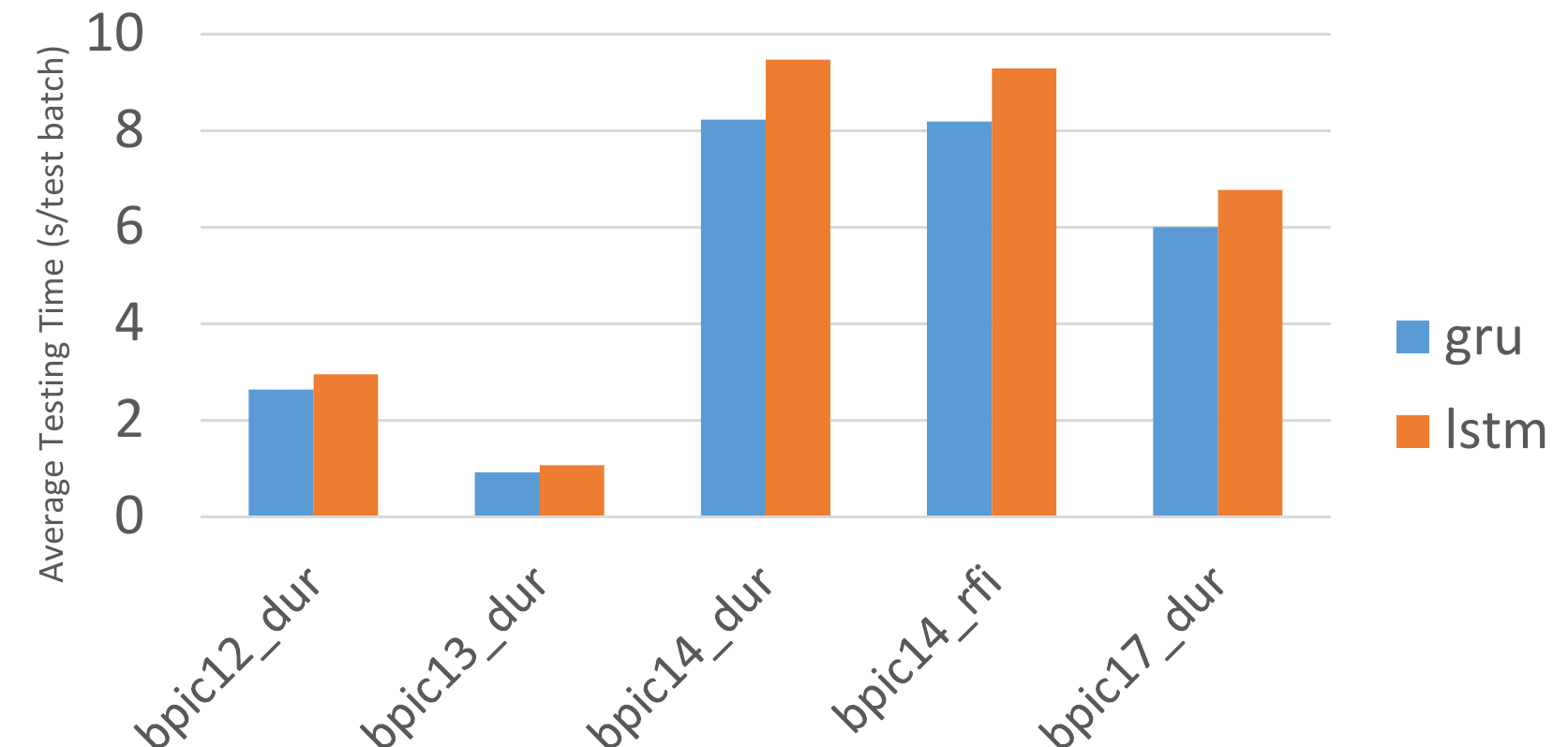}
				\caption[]{Testing time usage by RNN type}
				\label{figure:timeusage-rnntypes-test}
			\end{figure}
		\end{minipage}
	\end{minipage}
	\begin{minipage}{\linewidth}
		\centering
		\begin{minipage}{0.45\linewidth}
			\begin{figure}[H]
				\includegraphics[trim={2cm 10.0cm 2cm 11.6cm},clip,width=0.9\linewidth]{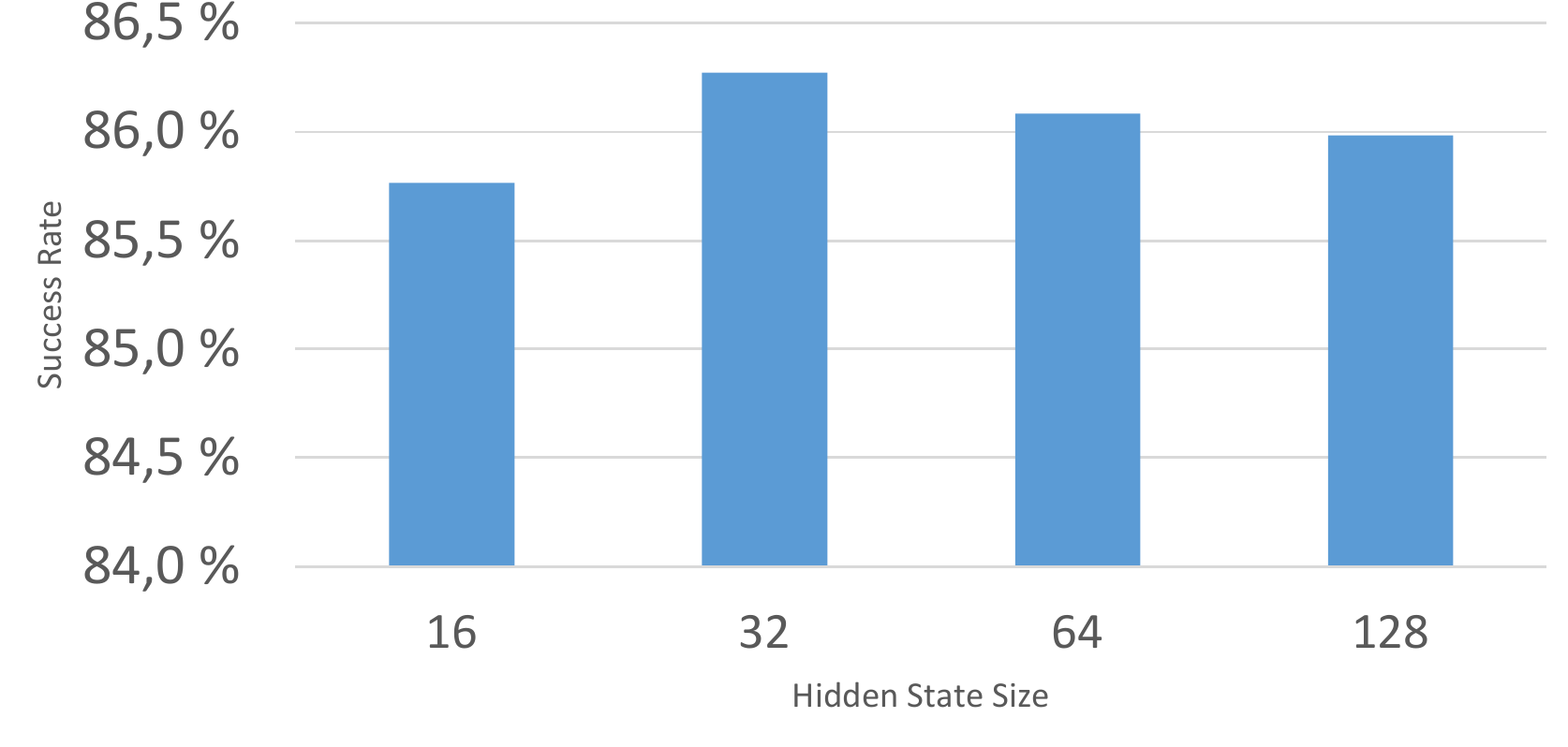}
				\caption[]{The effect of hidden state size to test accuracy}
				\label{figure:avgtest-byhiddensize}
			\end{figure}
		\end{minipage}
		\hspace{0.05\linewidth}
		\begin{minipage}{0.45\linewidth}
			\begin{figure}[H]
				\includegraphics[trim={2cm 10.0cm 2cm 11.6cm},clip,width=0.9\linewidth]{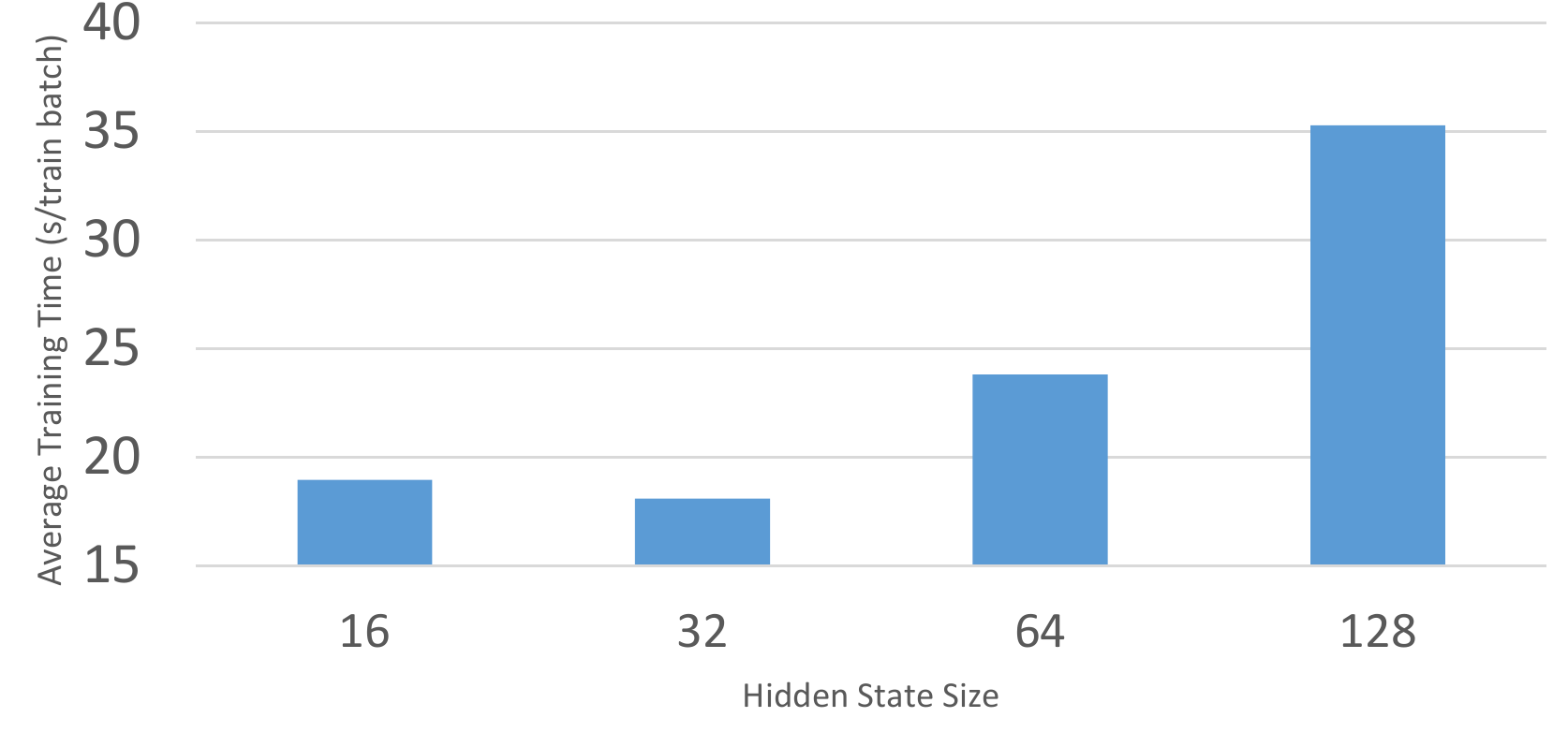}
				\caption[]{The effect of hidden state size to training time usage}
				\label{figure:avgtu-byhiddensize}
			\end{figure}
		\end{minipage}
	\end{minipage}
	\begin{minipage}{\linewidth}
		\centering
		\begin{minipage}{0.45\linewidth}
			\begin{figure}[H]
				\includegraphics[trim={2cm 10.0cm 2cm 11.4cm},clip,width=0.9\linewidth]{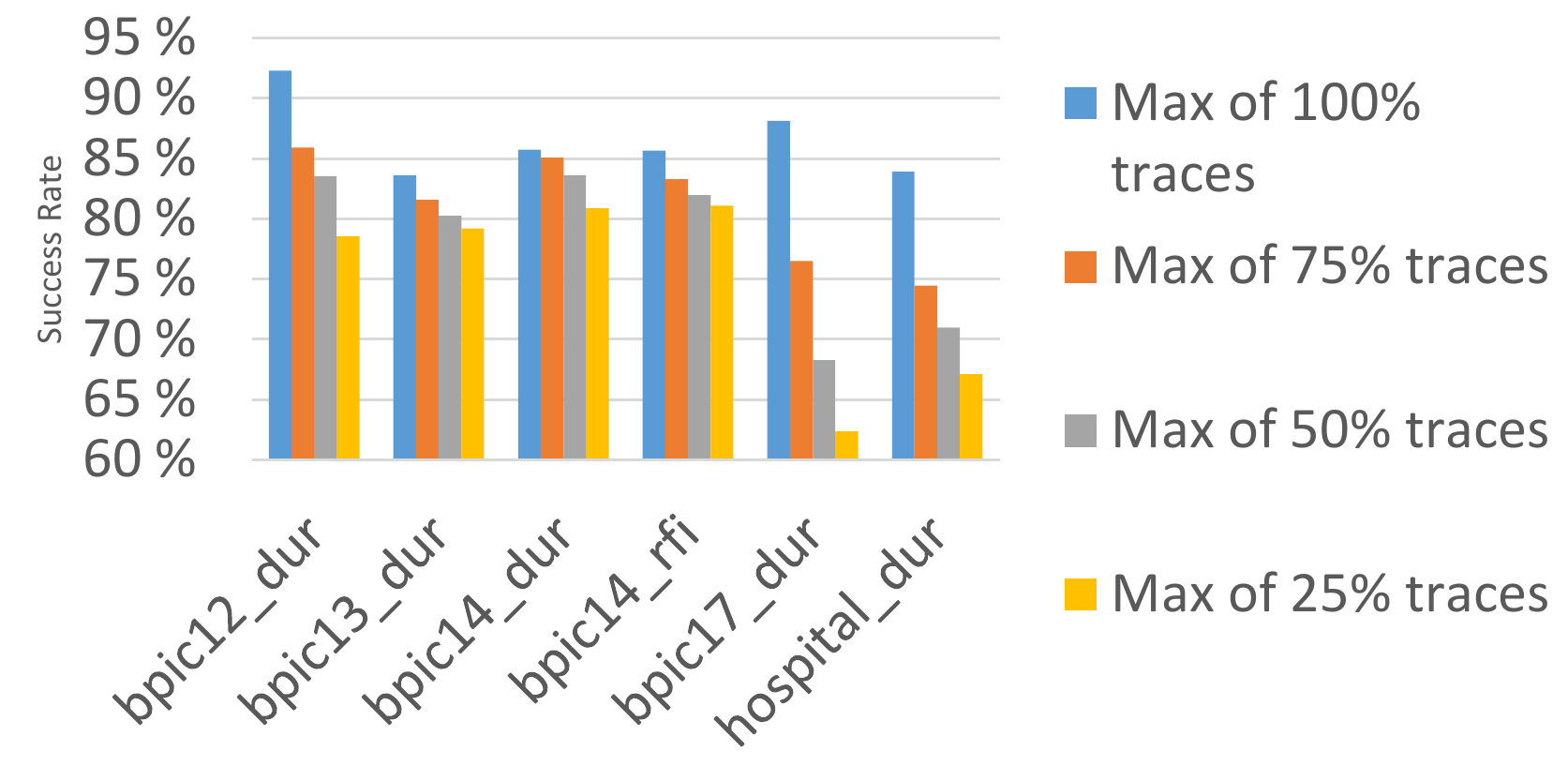}
				\caption[]{Prediction accuracy for incomplete traces}
				\label{figure:gru-allds-tt-maxpredacc-byds}
			\end{figure}
		\end{minipage}
		\hspace{0.05\linewidth}
		\begin{minipage}{0.45\linewidth}
			\begin{figure}[H]
				\includegraphics[trim={2cm 10.0cm 2cm 11.4cm},clip,width=0.9\linewidth]{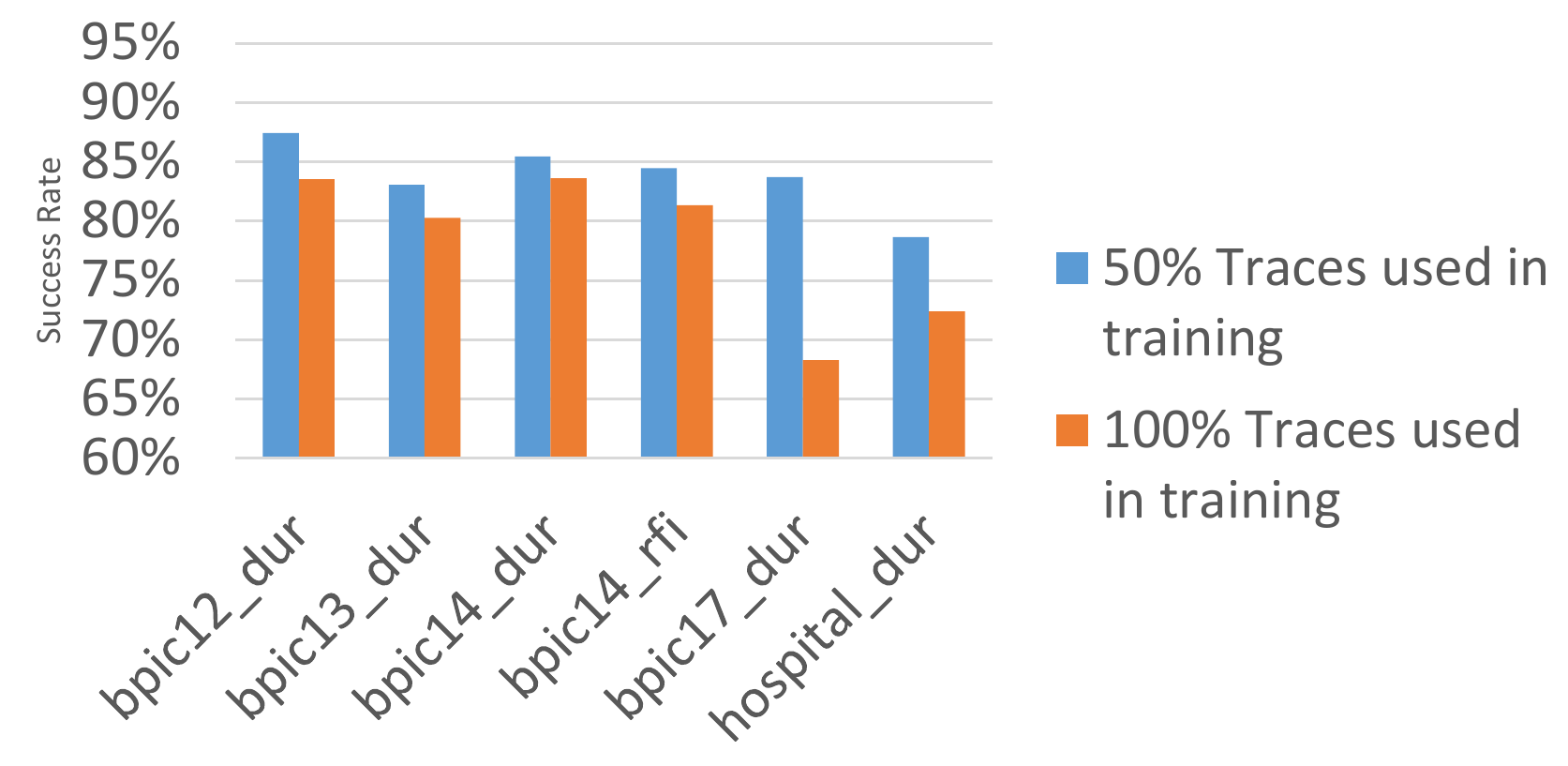}
				\caption[]{Prediction accuracy for incomplete traces using model trained using incomplete traces}
				\label{figure:gru-allds-50predacc-finalonly-trainedfor50vs100}
			\end{figure}
		\end{minipage}
	\end{minipage}
	\begin{minipage}{\linewidth}
		\centering
		\begin{minipage}{0.45\linewidth}
			\begin{figure}[H]
				\includegraphics[trim={2cm 10.0cm 2cm 11.4cm},clip,width=0.9\linewidth]{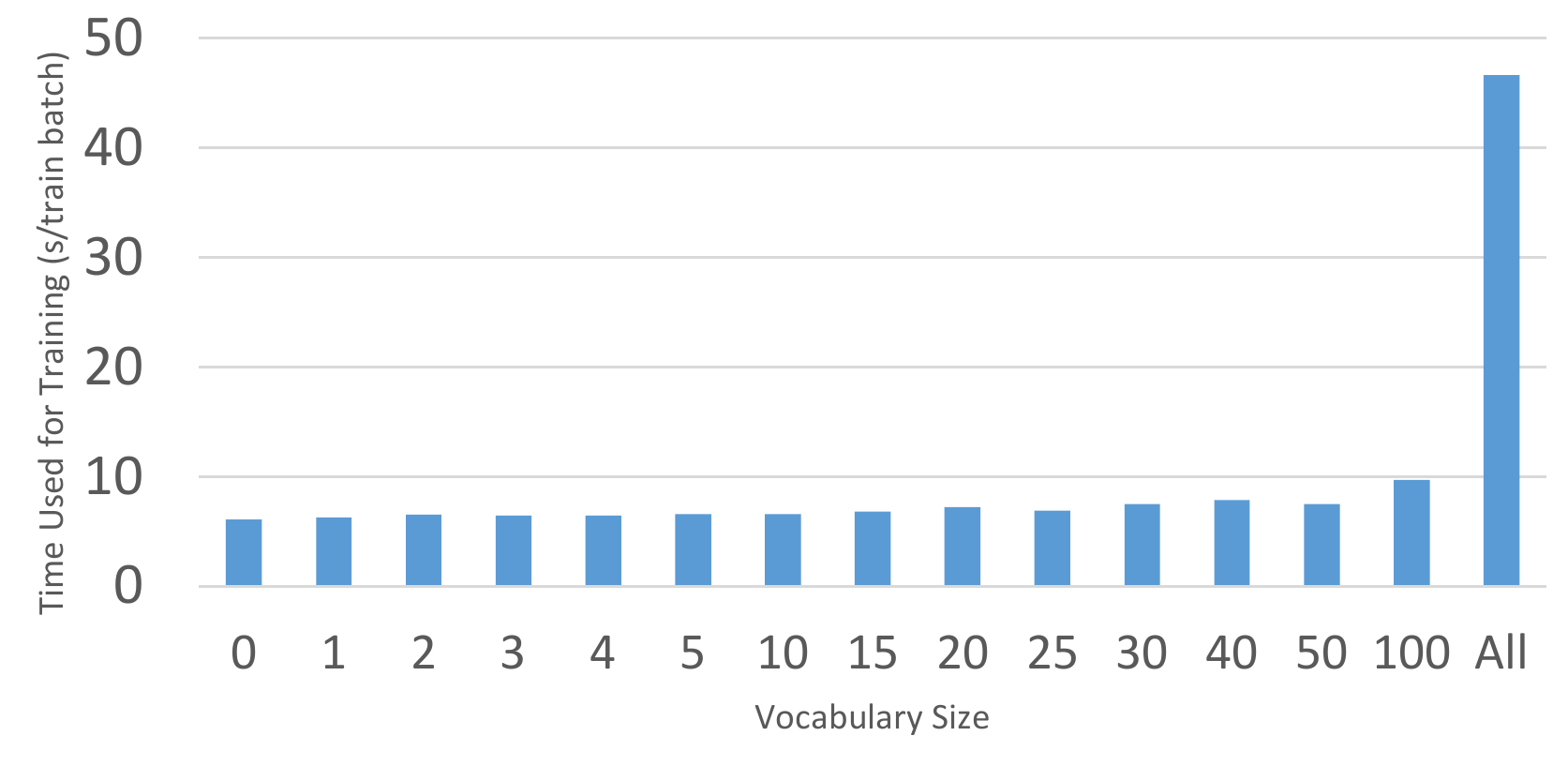}
				\caption[]{Training time usage by vocabulary size for Hospital dataset}
				\label{figure:gru-hosp-avgtu-byvocsize}
			\end{figure}
		\end{minipage}
		\hspace{0.05\linewidth}
		\begin{minipage}{0.45\linewidth}
			\begin{figure}[H]
				\includegraphics[trim={2cm 10.0cm 2cm 11.4cm},clip,width=0.9\linewidth]{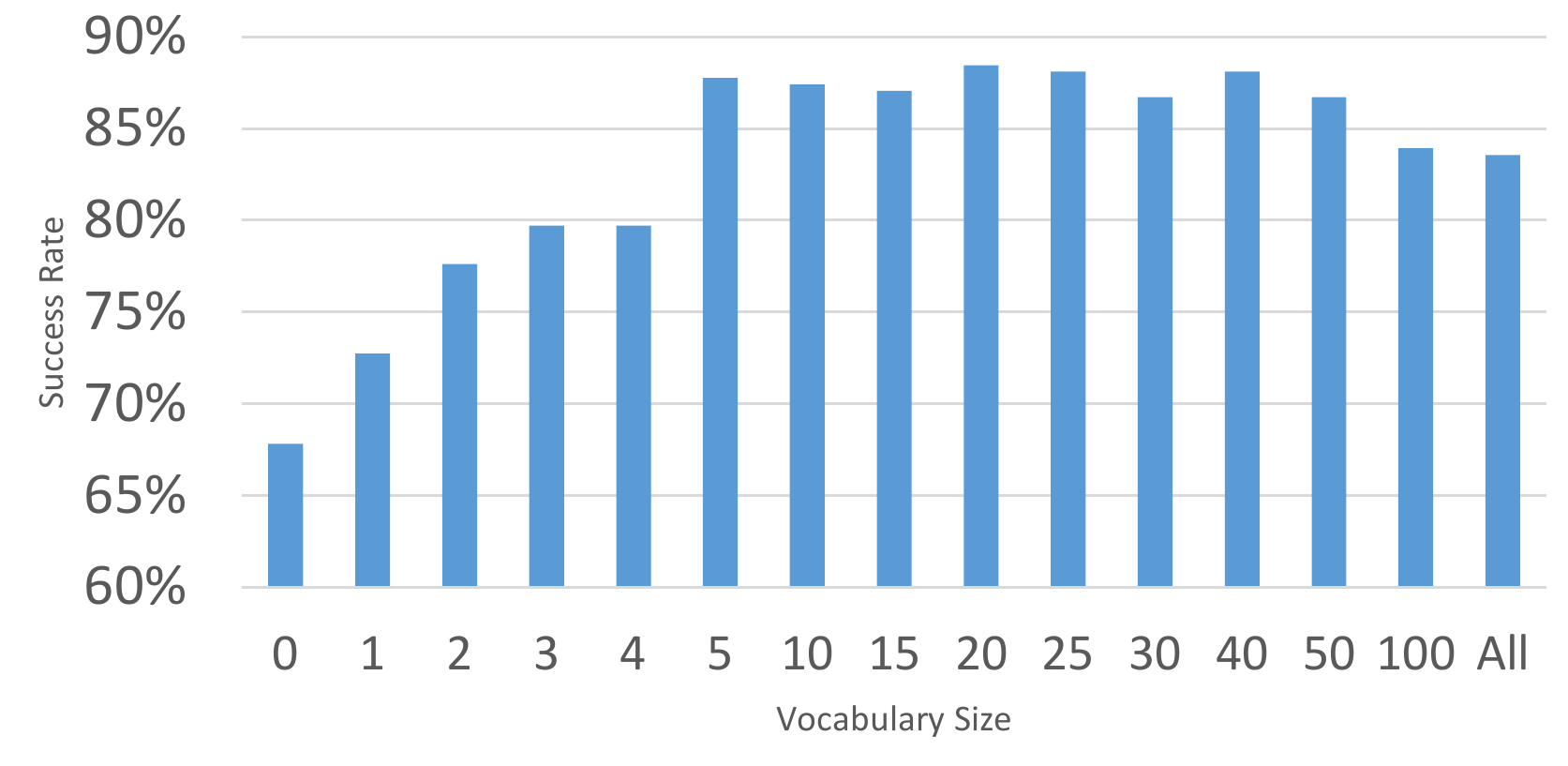}
				\caption[]{Maximum classification accuracy by vocabulary size for Hospital dataset}
				\label{figure:gru-hosp-maxacc-byvocsize}
			\end{figure}
		\end{minipage}
	\end{minipage}
\end{figure}

Figure~\ref{figure:accuracy-rnntypes} shows the maximum validation set classification accuracy results for all the tested datasets separately on both the RNN types having 32 as the size of the hidden state and the number of layers set to 1. Similarly Figure~\ref{figure:auroc-rnntypes} shows the AUROC values in the same test runs. From these results, it can be seen that LSTM and GRU both manage to get almost the same classification accuracies in both the measurements. AUROC values indicate that both RNN types have been created in a way that the model is able to classify data quite accurately and also that the model is not a trivial one such as always predicting a certain classification.

Next we measured the time usage when training and testing models. Figure~\ref{figure:timeusage-rnntypes} shows how long it took to train the model for total of 100000 traces. Similarly Figure~\ref{figure:timeusage-rnntypes-test} illustrates the average time usage for one test iteration in the test involving running the predictions four times for all the validation data set traces with all the tested activity sequence subsets. From these figures it can clearly be seen that GRU is faster to train and perform classifications with, than LSTM with similar hidden state sizes and the default initializations. Based on these results, we decided to use only GRU in our further tests.

The next step was to figure out whether it is of any use to use more than one GRU layer for our classification task. Based on our tests, it was found out that having two layers brings only very minimal value in our test case. In some of the tested datasets, it takes longer for the two layer model to even start getting any real advantage over the always predicting the most common outcome, whereas the one layer model learns clearly faster. In addition to this, it was seen that training two layers required double the amount of time. While the maximum accuracy was in some cases slightly better for the two layer model, we chose to continue our tests only with one layer model. It is also characteristic of the test runs that the accuracy first rises from the trivial classification accuracy to its maximum, after which it starts to slowly degrade. This indicates that after certain point, the model starts to over-fit the data and does not generalize that well any more.


Next, we wanted to test the effect of the hidden state size into the accuracy of the classification. Figure~\ref{figure:avgtest-byhiddensize} illustrates this by showing the average achieved accuracy for all the tested datasets, except Hospital. It can be seen that using hidden state size of 32 yields the most accurate results in the experimented cases. Since also Figure~\ref{figure:avgtu-byhiddensize} shows that the average time usage for each training iteration is the smallest when using 32 as the hidden state size, we chose 32 as the hidden state size for all the remaining experiments.

Next task we wanted to experiment with was the prediction: What is the accuracy of trace classification when the trace is still ongoing? Figure~\ref{figure:gru-allds-tt-maxpredacc-byds} shows the maximum prediction accuracies for all the experimented datasets when the model was trained only with full length traces but the validation was performed with continuous subsequences of 25\%, 50\%, 75\% and 100\% of all the activities within traces. Based on this figure, one can draw a conclusion that the prediction accuracy clearly depends on the data set and the classification task being performed. E.g., in BPIC12, BPIC13 and BPIC14 data set, it is possible to achieve over 80\% accuracy when the classification is performed based on the duration of the cases even when the trace being predicted has only 50\% of the activities of a full trace. However, BPIC17 and Hospital perform much worse both providing over 80\% accuracy only when given full traces. 

The next Figure~\ref{figure:gru-allds-50predacc-finalonly-trainedfor50vs100} compares the classification prediction accuracies of models built using two different methods. In the first method, the model is trained with full traces as in Figure~\ref{figure:gru-allds-tt-maxpredacc-byds} whereas in the second method the model is trained with first 50\% of the activities in the traces. In both the cases, 50\% traces are used as test data set. Based on these results, a conclusion can be drawn that it is best to train the model using traces having as similar characteristics as possible to the traces used in the testing. Thus, predicting the labeling of an ongoing trace, it is recommended that the model has also been trained with ongoing traces in similar phase. The phase could be measured, e.g., by measuring a life-time of the trace thus far or even using a separate neural network model trained for that purpose. It should, however, be noted also that for some datasets the prediction works nearly as well with full traces as with partial traces. Next we compared the maximum prediction accuracies achieved with GRUs to those achieved using \textit{Gradient Boosting Machine} (GBM)-based technique while also applying feature selection as discussed in paper~\cite{DBLP:conf/bpm/HinkkaLHJ17}. The differences in the accuracies achieved in these experiments are quite consistently in favor of GRU technique. Especially in the Hospital data set the accuracy improvement was exceptionally good.

We also compared the time required for GBM to reach its maximum accuracy for each dataset, in the experiments made for ~\cite{DBLP:journals/corr/abs-1710-02823}, with the time required for GRU to train a model that has at least similar accuracy as the GBM. In this test, GBM had better response times in BPIC12 and BPIC17. Hospital training time performance was also clearly worse in GRU. Partially the reason for that was the fact that we had to use four times smaller batch size in training since the GPU in the test system did not have enough memory to use larger batch sizes used in other datasets. Another issue to be noted especially in Hospital dataset is that by using feature selection the amount of features can be brought down to a very small number, for which GBM can be performed very efficiently. However, the GRU still has to work with full activity sequences and full vocabularies. 

For these purposes, we also experimented with vocabularies that were limited to a selected number of the most common activity identifiers. The results of these tests for Hospital data set are shown in Figure~\ref{figure:gru-hosp-avgtu-byvocsize} and Figure~\ref{figure:gru-hosp-maxacc-byvocsize}, which illustrate that the best training times as well as accuracies were achieved using a limited vocabulary. The time required to build the model with the vocabulary size of 20, which achieved the most accurate classification results, is only about 16\% of the time required to train with full vocabulary. Using this same model, the time required to reach GRU's best performance was 119 seconds, which still was slower than GBM's 29 seconds. Thus it seems that the length of the sequence is also a bottleneck. Finally, we made a test runs where vocabulary size was set to 20 and we also truncated all the sequences so that successive infrequent words that were not in vocabulary were replaced with only one occurrence of the word representing an infrequent word. In this case, the model training took about 20\% less time with very small effect to the accuracy. This way, we managed to reach the GBM's best performance in 46 seconds. Finally using similar approach for a model built using vocabulary of size 5, the result was achieved faster than GBM. In this case, the maximum sequence length was almost halved due to the truncation down to 610 unique activity identifiers. 

Thus, in the end, GRU managed to outperform more traditional GBM in all the measured metrics based on the classification accuracy and training time. GRU, which has not been earlier used in process mining context, also clearly outperformed LSTM in the required training time while still achieving similar accuracy. GRU based solutions offer also various other simple means to improve the accuracy and required training time, such as using different gradient descent optimization algorithms, modifying the learning rate and also using bigger batch sizes if there is enough memory available in the GPU. Also, in order to avoid overfitting, a regularization method, such as dropout, could be applied.

\section{Conclusions}
\label{conclusions}

Employing Recurrent Neural Network based classification for process mining traces, processed by Natural Language Processing techniques into sequences of \textit{words}, can achieve at least similar level of performance as feature selection and GBM based classification. One big advantage for RNN based solution is that the amount of input data required is very small; just the list of traces with their activity sequences and the classification information for the training data. For more traditional classification solutions, user needs to provide the full set of features based on which the training and classifications are made. Extracting these features out of activity sequences can be a very expensive process in itself. This feature selection was the core of our earlier paper~\cite{DBLP:conf/bpm/HinkkaLHJ17}. Our experiments in this paper clearly showed that GRU based models yield more accurate classification results faster than more traditional GBM based classification using any of the structural feature combinations experimented in our previous work. All the experiments were performed on a framework that offloaded most of the calculations to GPU for improved performance and scalability. 

One of our main results is the suggestion to use GRU models for predicting process instance outcomes as GRU, that has not previously been used in process mining context, is usually better choice for RNN type than LSTM mostly due to it being faster to train and its ability to achieve almost identical classification and prediction accuracy. We experimented also two approaches for the prediction of eventual classification label for still ongoing traces. From this test we found out that it is always clearly better to train a model with traces at as similar phases of their lifetime as possible to the traces being tested.

We also investigated how to improve the required training time, especially when using data sets having long activity sequences and a lot of activities compared to the number of training data activity sequences. We found out that the number of activities can be decreased by treating all the infrequent activities as one activity without it having a big effect to the classification accuracy, while still having a noticeable effect in throughput time and GPU memory requirements. We also found out that replacing long sequences of infrequent activities with just one activity representing all the infrequent activities can further improve the throughput time without it affecting dramatically into classification accuracy.

All the raw test results gathered from the performed experiments, some of which was not discussed nor explored in this paper in detail, together with the developed python source code for the test framework, can be found in the support materials~\cite{supportmaterials}. 



\section{Acknowledgements}
\label{acknowledgements}
We want to thank QPR Software Plc for funding our research. Financial support of Academy of Finland projects 139402 and 277522 is acknowledged.

\label{references}
\bibliography{paper}

\end{document}